%% file: main.tex
\documentclass[sigconf]{aamas}

\usepackage{graphics}
\usepackage{multirow}
\usepackage{algorithm}
\usepackage{subcaption}

\allowdisplaybreaks
\newcommand{\etal}{\textit{et al.}}

\makeatletter
\gdef\@copyrightpermission{
  \begin{minipage}{0.2\columnwidth}
   \href{https://creativecommons.org/licenses/by/4.0/}{\includegraphics[width=0.90\textwidth]{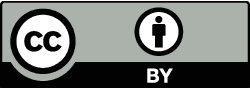}}
  \end{minipage}\hfill
  \begin{minipage}{0.8\columnwidth}
   \href{https://creativecommons.org/licenses/by/4.0/}{This work is licensed under a Creative Commons Attribution International 4.0 License.}
  \end{minipage}
  \vspace{5pt}
}
\makeatother

\setcopyright{ifaamas}
\acmConference[AAMAS '25]{Proc.\@ of the 24th International Conference
on Autonomous Agents and Multiagent Systems (AAMAS 2025)}{May 19 -- 23, 2025}
{Detroit, Michigan, USA}{Y.~Vorobeychik, S.~Das, A.~Nowé  (eds.)}
\copyrightyear{2025}
\acmYear{2025}
\acmDOI{}
\acmPrice{}
\acmISBN{}

\definecolor{White}{rgb}{1,1,1}
\definecolor{LightBlue}{rgb}{0.8235,0.9737,0.9882}
\definecolor{LightGray}{HTML}{ececec}
\definecolor{LightGreen}{HTML}{effdee} % e4fde1
\definecolor{LightYellow}{HTML}{fffee0} % fffec8, fffdaf
\definecolor{LightOrange}{HTML}{fee7da} % fee2d2
\definecolor{textGreen}{HTML}{008000}
\definecolor{textRed}{HTML}{cc0000}
\newcommand{\redtext}[1]{\textcolor{red}{#1}}
\newcommand{\bluetext}[1]{\textcolor{blue}{#1}}
\newcommand{\greentext}[1]{\textcolor{green}{#1}}

\newcommand{\myparagraph}[1]{\vspace{4pt}\noindent{\bf #1}}

\acmSubmissionID{778}

\title[AAMAS-2025 Formatting Instructions]{GUIDE-CoT: Goal-driven and User-Informed Dynamic Estimation for Pedestrian Trajectory using Chain-of-Thought}

\author{Sungsik Kim}
\affiliation{
  \institution{Kookmin University}
  \city{Seoul}
  \country{Korea}}

\author{Janghyun Baek}
\affiliation{
  \institution{Korea University}
  \city{Seoul}
  \country{Korea}}

  \author{Jinkyu Kim}
\affiliation{
  \institution{Korea University}
  \city{Seoul}
  \country{Korea}}
  \email{jinkyukim@korea.ac.kr}

\author{Jaekoo Lee}
\affiliation{
  \institution{Kookmin University}
  \city{Seoul}
  \country{Korea}}
  \email{jaekoo@kookmin.ac.kr}

\input{tex/0_abstract}


\newcommand{\BibTeX}{\rm B\kern-.05em{\sc i\kern-.025em b}\kern-.08em\TeX}

\begin{document}
\sloppy

%%% The following commands remove the headers in your paper. For final 
%%% papers, these will be inserted during the pagination process.

\pagestyle{fancy}
\fancyhead{}

%%% The next command prints the information defined in the preamble.

\maketitle 

%%%%%%%%%%%%%%%%%%%%%%%%%%%%%%%%%%%%%%%%%%%%%%%%%%%%%%%%%%%%%%%%%%%%%%%%

\input{tex/1_introduction}
\input{tex/2_related_works}
\input{tex/3_method}

\input{tex/4_experiments}
\input{tex/5_conclusion}

%%%%%%%%%%%%%%%%%%%%%%%%%%%%%%%%%%%%%%%%%%%%%%%%%%%%%%%%%%%%%%%%%%%%%%%%

%%% The acknowledgments section is defined using the "acks" environment
%%% (rather than an unnumbered section). The use of this environment 
%%% ensures the proper identification of the section in the article 
%%% metadata as well as the consistent spelling of the heading.

\begin{acks}
This research was supported by Institute of Information \& Communications Technology Planning \& Evaluation(IITP)-ITRC(IITP-2025-RS-2022-00156295; Information Technology Research Center), the grant (IITP-2024-RS-2024-00397085; Leading Generative AI Human Resources Development, and IITP-2024-RS-2024-00417958; Global Research Support
Program in the Digital Field program), and the National Research
Foundation of Korea(NRF) grant (No.RS-2023-00212484; xAI for Motion
Prediction in Complex, Real-World Driving Environment) funded by the Korea government(MSIT)
% If you wish to include any acknowledgments in your paper (e.g., to 
% people or funding agencies), please do so using the `\texttt{acks}' 
% environment. Note that the text of your acknowledgments will be omitted
% if you compile your document with the `\texttt{anonymous}' option.
\end{acks}

%%%%%%%%%%%%%%%%%%%%%%%%%%%%%%%%%%%%%%%%%%%%%%%%%%%%%%%%%%%%%%%%%%%%%%%%

%%% The next two lines define, first, the bibliography style to be 
%%% applied, and, second, the bibliography file to be used.

\bibliographystyle{ACM-Reference-Format} 
\bibliography{main}

%%%%%%%%%%%%%%%%%%%%%%%%%%%%%%%%%%%%%%%%%%%%%%%%%%%%%%%%%%%%%%%%%%%%%%%%

\end{document}

%% file: tex/0_abstract.tex
\begin{abstract}
While Large Language Models (LLMs) have recently shown impressive results in reasoning tasks, their application to pedestrian trajectory prediction remains challenging due to two key limitations: insufficient use of visual information and the difficulty of predicting entire trajectories. To address these challenges, we propose \textit{Goal-driven and User-Informed Dynamic Estimation for pedestrian trajectory using Chain-of-Thought (GUIDE-CoT)}. Our approach integrates two innovative modules: (1) a goal-oriented visual prompt, which enhances goal prediction accuracy combining visual prompts with a pretrained visual encoder, and (2) a chain-of-thought (CoT) LLM for trajectory generation, which generates realistic trajectories toward the predicted goal. Moreover, our method introduces controllable trajectory generation, allowing for flexible and user-guided modifications to the predicted paths. Through extensive experiments on the ETH/UCY benchmark datasets, our method achieves state-of-the-art performance, delivering both high accuracy and greater adaptability in pedestrian trajectory prediction. Our code is publicly available at \url{https://github.com/ai-kmu/GUIDE-CoT}.
\end{abstract}

%% file: tex/1_introduction.tex
\begin{figure}[t]
    \centering
    \begin{subfigure}[b]{.48\textwidth}
        \centering
        \includegraphics[width=\textwidth]{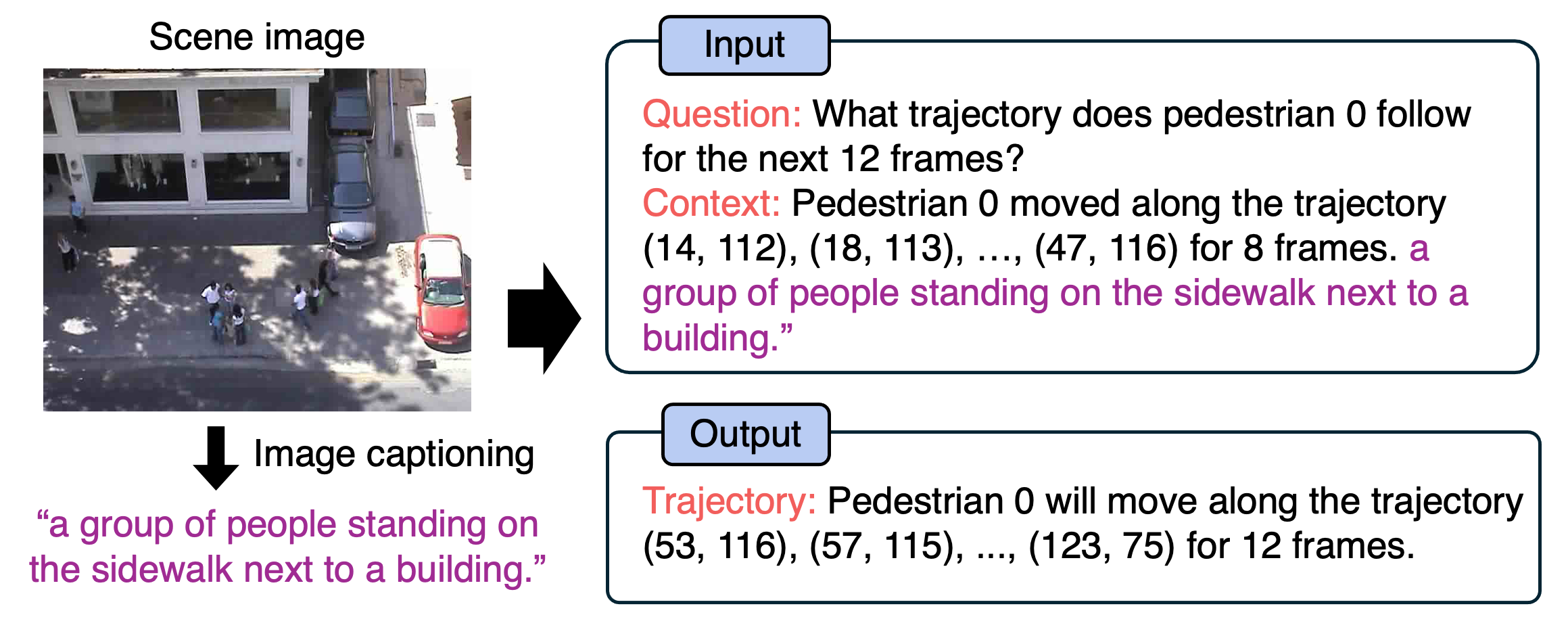}
        \caption{
        Conventional LLM-based approaches
        }
        \label{fig:1a}
    \end{subfigure}
    \hfill
    \begin{subfigure}[b]{.475\textwidth}
        \centering
        \includegraphics[width=\textwidth]{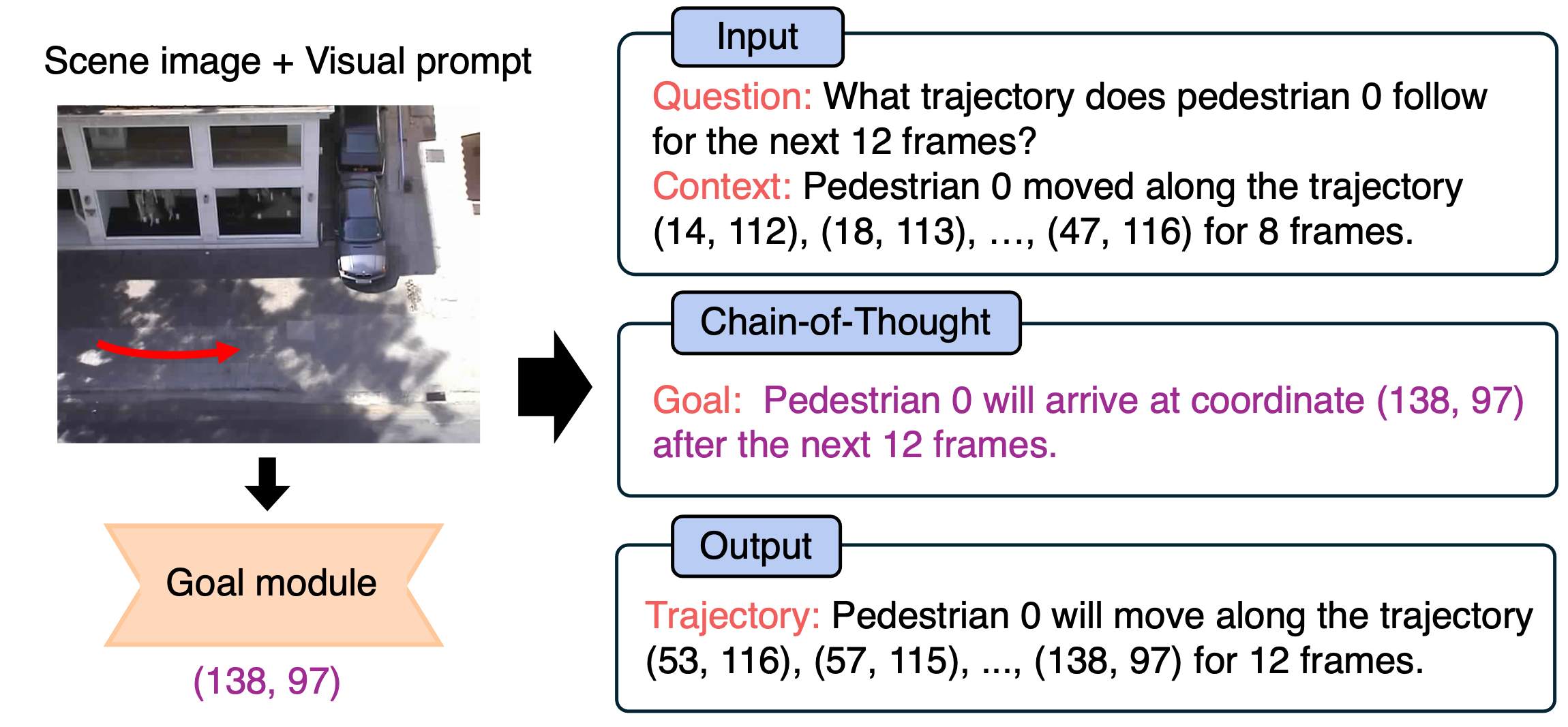}
        \caption{
        Our proposed approach (GUIDE-CoT)
        }
        \label{fig:1b}
    \end{subfigure}
    \caption{
    Comparison between conventional LLM-based methods and our approach for pedestrian trajectory prediction. (a) Conventional approaches often leverage LLM's reasoning capability to predict pedestrians' future trajectories conditioned on textual contexts, which contain their past observations and scene descriptions (from an off-the-shelf image captioning model). (b) Our proposed method, called GUIDE-CoT, further improves the model's prediction performance by predicting pedestrians' final goal position given the scene image with overlaid visual prompts (i.e., a red arrow). Such predicted goal position is then augmented into an LLM in a similar manner to Chain-of-Thought (CoT), offering rich intermediate reasoning contexts. 
    }
    \label{fig:teaser}
\end{figure}

\section{Introduction}
Pedestrian trajectory prediction has emerged as a critical task in various applications, including autonomous driving and urban planning~\cite{trace_and_pace}. In autonomous driving, vehicles or robots must dynamically adjust their paths by predicting and responding to the movements of nearby pedestrians~\cite{interaction}. The inability of an autonomous agent to accurately predict future pedestrian trajectories can lead to severe consequences, such as collisions, posing significant safety risks~\cite{social_gan}. Furthermore, in the context of urban planning, particularly in smart cities, accurately understanding and predicting pedestrian flow is crucial. This insight enables the optimization of transportation networks and infrastructure, leading to smoother traffic management and greater overall efficiency~\cite{smart_city}.

Deep learning has led to substantial breakthroughs in pedestrian trajectory prediction, with recent studies showing that deep learning models trained on large-scale datasets can accurately predict a wide range of pedestrian behaviors~\cite{pedtp_review}. While existing approaches have focused on modeling pedestrian interactions and environmental factors, there has been growing interest in goal-based pedestrian trajectory prediction~\cite{pecnet, ynet, goal_gan, goal_sar}. These methods aim to predict a pedestrian's future destination based on their past trajectory and map information. 
Since the accurate prediction of a pedestrian's goal is crucial for trajectory estimation, improving goal prediction accuracy has become a central focus in trajectory prediction research.

In this work, we propose a novel approach by leveraging large language models (LLMs), which have recently been applied to a variety of reasoning and predicting tasks across domains~\cite{cot, llm_science, llm_medical_reasoning}. Building on prior work, which has shown that converting a pedestrian's past trajectory into a natural language format and feeding it into an LLM can result in state-of-the-art (SOTA) trajectory prediction performance~\cite{lmtrajectory}, we propose a method that significantly enhances LLM-based methods by incorporating goal prediction as a key step in the chain-of-thought (CoT) reasoning process. As illustrated in Figure~\ref{fig:teaser}, unlike existing LLM-based methods that predict trajectories directly from historical trajectories and scene image caption, our approach first predicts CoT reasoning of the pedestrian’s goal to guide the final trajectory generation.

In contrast to existing methods that rely on semantic segmentation maps with trajectory heatmaps~\cite{ynet, goal_sar} or dynamic features~\cite{goal_gan} for goal prediction, our approach introduces a goal-oriented visual prompt that leverages visual cues more effectively.
To be specific, we enhance goal prediction by incorporating visual prompts, represented as directional arrows on RGB images to indicate pedestrian movement. These visual prompts are processed by a pretrained model, and rather than solely depending on the model’s output for direct goal prediction, we combine the visual features with semantic map-based goal prediction. This approach enables more accurate and context-aware predictions by utilizing both spatial and visual information. The predicted goals are then integrated into the CoT reasoning process, which provides context for the LLM, guiding it to generate more precise future trajectories through structured reasoning.

We propose a controllable trajectory generation method by manipulating the CoT reasoning process. 
Specifically, we explore how modifying the goal context during inference can adjust the predicted trajectory, enabling user-guided control over future path predictions. 
Experiments on the ETH/UCY benchmark datasets show that our method delivers results competitive with state-of-the-art approaches, while introducing new capabilities in controllability and reasoning. 

Our main contributions are summarized as follows:
\begin{itemize}
    \item We introduce a goal-oriented visual prompt that effectively integrates spatial and visual information for superior goal prediction, enhancing trajectory prediction performance.
    \item We propose a CoT reasoning prompt that leverages goal predictions as intermediate steps, enabling the LLM to predict future trajectories more accurately. 
    \item We enable controllable trajectory generation, allowing users to modify goal context during inference, providing dynamic control over the predicted trajectories.
\end{itemize}

%% file: tex/2_related_works.tex
\section{Related Work}

\begin{figure*}[t]
    \centering
    \begin{subfigure}[b]{.99\textwidth}
        \centering
        \includegraphics[width=\textwidth, trim=7 2 1 6, clip]{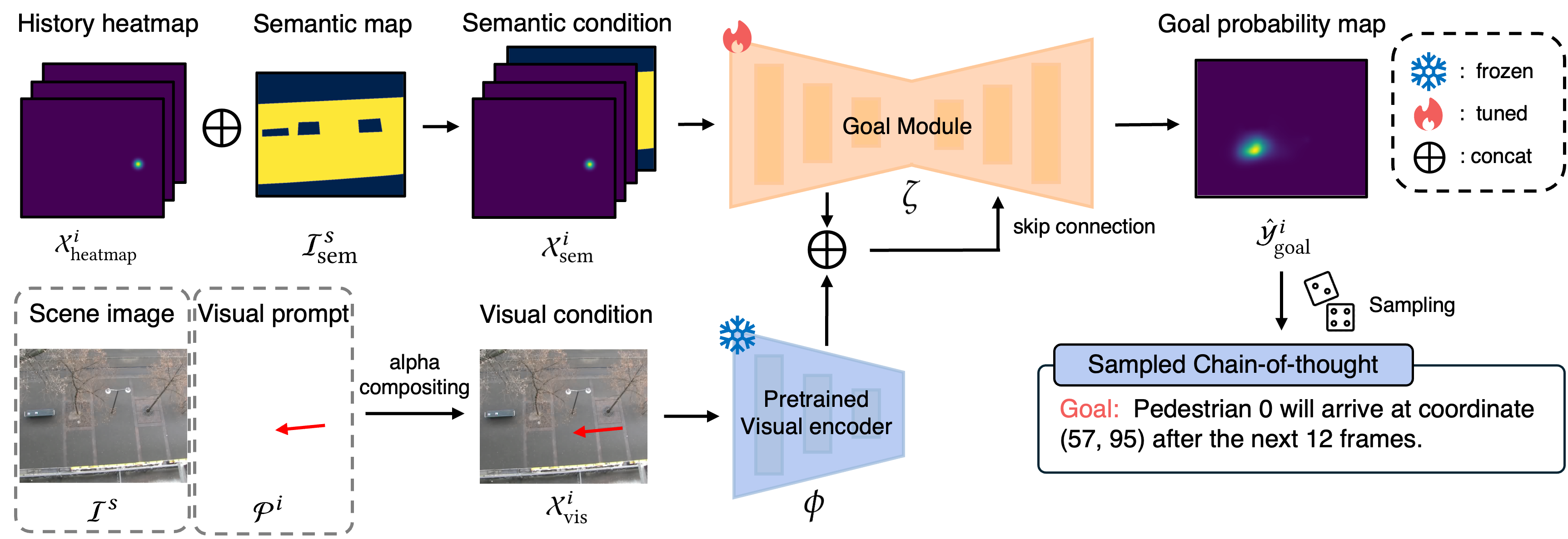}
        \caption{
        Goal predictor with visual prompt
        }
        \label{fig:2a}
    \end{subfigure}
    \hfill
    \begin{subfigure}[b]{.99\textwidth}
        \centering
        \includegraphics[width=\textwidth]{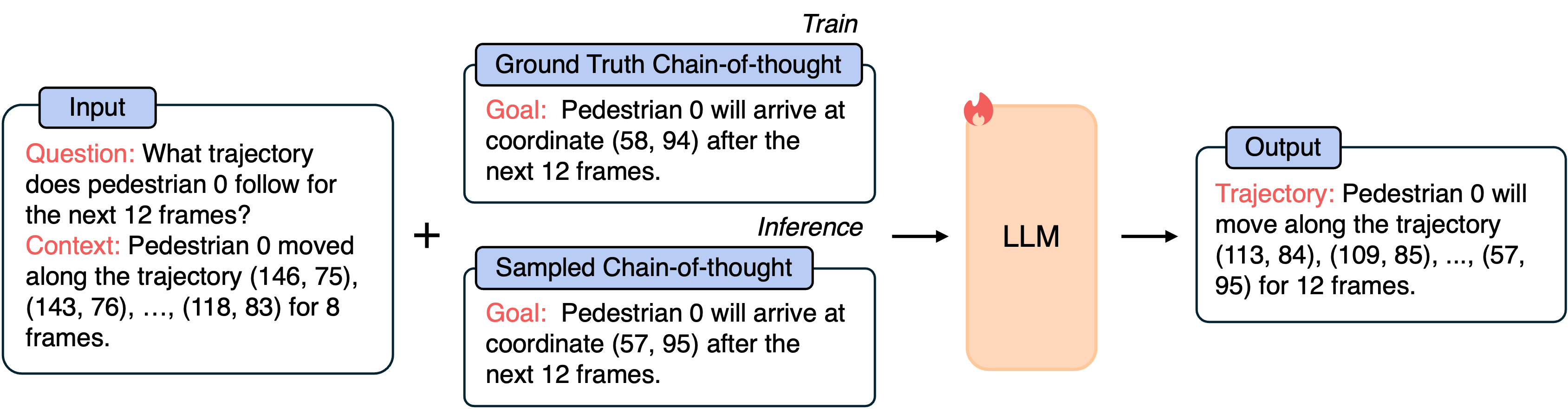}
        \caption{
        LLM-based trajectory prediction with Chain-of-Thought (CoT) reasoning
        }
        \label{fig:2b}
    \end{subfigure}
    \caption{
    An overview of our proposed approach, called GUIDE-CoT. (a) Our model first predicts each pedestrian's final goal position given (i) pedestrians' past observations, (ii) semantic BEV map, and (iii) top-down view scene image with visual prompt (i.e., a red arrow). Our model generates a sentence describing their final positions, such as ``Pedestrian 0 will arrive at coordinate (57, 95) after the next 12 frames.'' (b) Such a generated goal description is then augmented into the LLM in a similar way to Chain-of-Thought reasoning, generating the final trajectory of each pedestrian.}
    \label{fig:main_framework}
\end{figure*}

\subsection{Pedestrian Trajectory Prediction}
Pedestrian trajectory prediction has been extensively studied, with early approaches relying on simple physics-based models~\cite{social_force_model}. As deep learning advanced, data-driven models that learn pedestrian movement patterns and predict future trajectories become the dominant methodology. A notable example is Social-LSTM~\cite{social_lstm}, which introduced LSTM-based modeling for pedestrian paths. Following this, a wide variety of learning approaches, including GANs~\cite{social_gan, sophie, social_ways, mg-gan}, CVAEs~\cite{trajectron++, socialvae}, Diffusion models~\cite{mid, led, dice}, GCNs~\cite{social_stgcnn, semantics-stgcnn, d-stgcn}, and Transformers~\cite{star, agentformer}, have been explored to further improve prediction accuracy.

While these approaches have made significant strides, many focus solely on predicting the immediate future path based on historical data and interactions with the environment. However, goal-based pedestrian trajectory prediction has recently emerged as an important alternative. This approach assumes that pedestrian trajectories are heavily influenced by their intended destination, rather than just local interactions. For instance, Y-Net~\cite{ynet} leverages history trajectories and environmental context (in the form of heatmaps and semantic segmentation map) and uses U-Net~\cite{unet} to predict a goal probability map. This goal-based approach is critical because predicting the final destination can significantly enhance trajectory accuracy.

Most existing methods still struggle to effectively incorporate information of multiple modalities and handle the complexity of trajectory prediction as a reasoning process. Recent advancements in LLMs have demonstrated strong reasoning capabilities across various domains, including pedestrian trajectory prediction. As a LLM-based SOTA model, the LMTraj model~\cite{lmtrajectory} reframed trajectory prediction as a question-answering task, utilizing LLMs to infer pedestrian movements, and achieved superior performance over alternatives. However, LLM-based models~\cite{lmtrajectory} have two key limitations: insufficient integration of visual information and difficulty in predicting the full trajectory. 

\subsection{Large Language Models for Sequence Prediction}
Large Language Models (LLMs) have driven significant advances in natural language processing by leveraging large datasets and transformer architectures.
These models have achieved impressive results across a wide range of applications, such as text generation, translation, and summarization~\cite{bert, gpt-2, t5}. 
Larger LLMs~\cite{gpt-3, gpt4} have shown an enhanced ability to generalize to unseen data and tasks, demonstrating robust performance even in few-shot and zero-shot learning scenarios, where limited task-specific examples are available.

LLMs have been widely utilized in various sequence prediction tasks to enhance inference performance. For example, Time-LLM~\cite{timellm} leverages a pretrained LLM for time series forecasting and has demonstrated high performance in both zero-shot and few-shot settings. Furthermore, recent studies have proposed the use of LLMs to predict complex biological sequences~\cite{tfcl}, such as DNA patterns~\cite{dnagpt} and protein structures~\cite{protgpt2}. Specifically, DNAGPT~\cite{dnagpt} introduced a generalized pretrained LLM for modeling and predicting DNA sequences, while ProtGPT2~\cite{protgpt2} presented a GPT-2~\cite{gpt-2}–based protein sequence generation model that samples novel protein sequences. Similarly, studies have emerged exploring the use of LLMs to predict pedestrian trajectories--a sequential prediction task.

\subsection{Reasoning via Visual Prompt}
To enhance the reasoning capabilities of Large Multimodal Models (LMMs), additional visual information, known as visual prompts, has been introduced~\cite{vip-llava}. Visual prompts provide crucial visual cues that assist models in making more accurate inferences. One common approach is to integrate learnable parameters directly into the visual data. For instance, VPT~\cite{vpt} proposes adding learnable prompts to the input patches of each layer in the Vision Transformer~\cite{vit}. Another method involves padding image borders with learnable parameters embedding additional information around the edges of the image to guide the model’s attention~\cite{exploring_visual_prompt}. 

In addition, visual markers are often employed to highlight specific objects or regions within an image.
For example, CPT~\cite{cpt} uses unique colors to distinguish objects in region proposals, while Shtedritski~\etal~\cite{clip_red_circle} employs red circles to mark objects, leveraging CLIP~\cite{clip} for zero-shot reasoning. ViP-LLaVA~\cite{vip-llava} expands visual prompts by incorporating arrows, points, and triangles for more diverse reasoning tasks.

Inspired by these approaches, we propose a novel method to represent pedestrian trajectories using visual prompts. Our approach guides the trajectory of pedestrians, allowing the model to better understand movement patterns and predict future paths more effectively.

%% file: tex/3_method.tex
\section{Method}

\begin{figure}
    \centering
    \begin{subfigure}[b]{.46\textwidth}
        \centering
        \includegraphics[width=\textwidth]{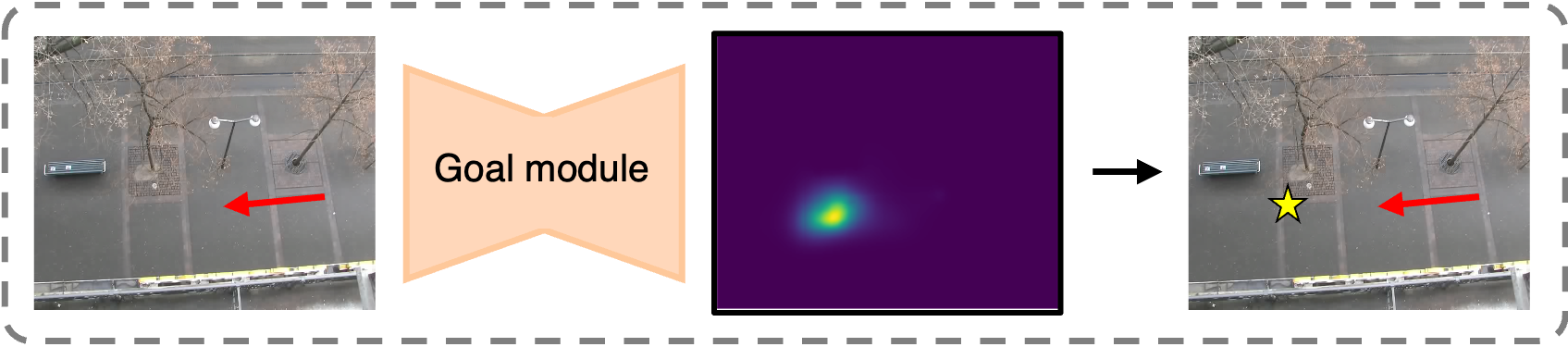}
        \caption{without user-provided guidance}
        \label{fig:3a}
    \end{subfigure}
    \hfill
    \begin{subfigure}[b]{.46\textwidth}
        \centering
        \includegraphics[width=\textwidth]{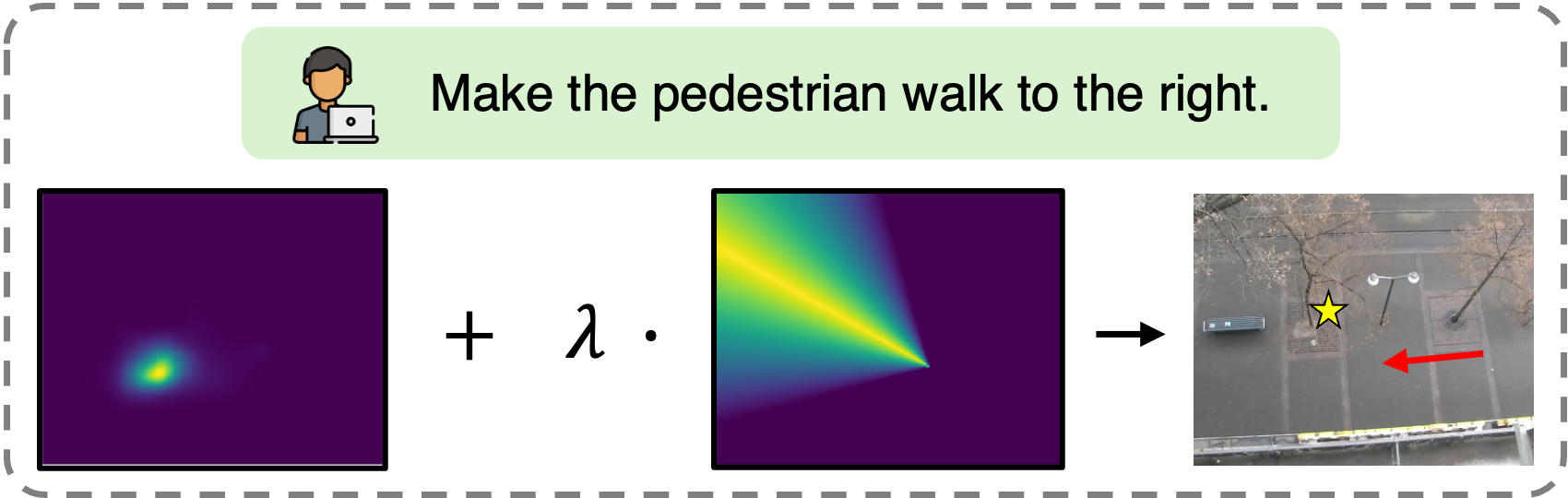}
        \caption{
        with directional user guidance
        }
        \label{fig:3b}
    \end{subfigure}
    \hfill
    \begin{subfigure}[b]{.46\textwidth}
        \centering
        \includegraphics[width=\textwidth]{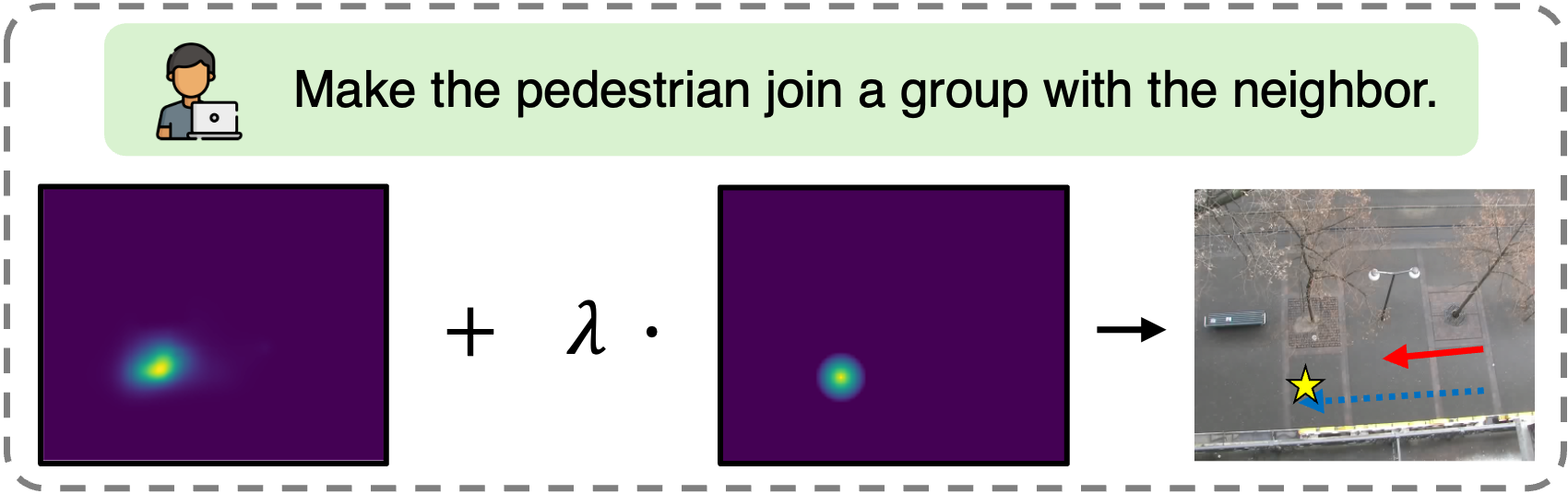}
        \caption{
        with positional user guidance
        }
        \label{fig:3c}
    \end{subfigure}
    \caption{
    Our goal-conditioned model further allows the user to provide text-driven guidance, controlling the model's trajectory prediction process. Such guidance may contain (b) a directional guide (e.g., ``make the pedestrian walk to the right'') or (c) a positional guide (e.g., ``make the pedestrian join a group with the neighbor''). Compare results with and without user-provided guidance, i.e., (a) vs. (b) and (c). 
    }
    \label{fig:control}
\end{figure}

\subsection{Problem Definition}
Pedestrian trajectory prediction task aims to forecast future trajectories of pedestrians based on their past movements. For each pedestrian $i$ at time step $t$, the observed past trajectory is composed of 2D position coordinates $p$ spanning a time window of length $\tau_{\text{obs}}$. This trajectory is represented as:
\begin{equation}
\mathcal{X}^{i}_{t} = \left\{ p_{t-\tau_{\text{obs}}+1}^i, \;\dots,\; p_t^i \right\}
\end{equation}
The input to the model consists of the past trajectories of $N$ pedestrians in the scene, denoted as:
\begin{equation}
\mathcal{X}_{t} = \left( \mathcal{X}^{0}_{t}, \;\dots,\; \mathcal{X}^{N-1}_{t} \right) \in \mathbb{R}^{N \times \tau_{\text{obs}} \times 2}
\end{equation}
Additionally, the model is provided with a scene image $\mathcal{I}^s$ and a corresponding semantic map $\mathcal{I}^s_{\text{sem}}$ for the scene $s$.

The goal of task is to predict the future trajectories of these $N$ pedestrians over a prediction window of length $\tau_{\text{pred}}$. The predicted future trajectories are represented as:
\begin{equation}
\hat{\mathcal{Y}}_{t} = \left(\mathcal{Y}^{0}_{t}, \;\dots,\; \mathcal{Y}^{N-1}_{t} \right) \in \mathbb{R}^{N \times \tau_{\text{pred}} \times 2}
\end{equation}
The objective of learning is to minimize the error between the predicted future trajectories $\hat{\mathcal{Y}}_t$ and the ground-truth future trajectories $\mathcal{Y}_t$.

\subsection{Goal-oriented Visual Prompt for Trajectory}
Building on prior work that demonstrates improvements in reasoning tasks through the visual knowledge of pretrained vision models~\cite{reasoning_survey}, we propose a novel approach to effectively leverage visual information from scene images for the goal prediction task.

A naïve approach might consider using top-view video frames of pedestrians for trajectory prediction via a video encoder. However, the naïve approach faces three key challenges: (1) obtaining top-view pedestrian videos is impractical, (2) distinguishing pedestrians within the scene is challenging, and (3) video processing is computationally inefficient compared to handling individual images.

To address these challenges while still utilizing the rich visual information from scene images, we introduce visual prompts—such as circles, arrows, and points~\cite{clip_red_circle, vip-llava}—to designate pedestrian positions and movements for goal prediction. Specifically, our approach uses arrow-shaped visual prompts to guide the goal prediction, allowing the model to better interpret and predict future trajectories.

As shown in Figure~\ref{fig:2a}, given the past trajectory $\mathcal{X}^i$ of the $i$-th pedestrian as input, we generate a visual prompt $\mathcal{P}^i = Draw \big( \mathcal{X}^i \big)$ by connecting the 2D coordinates with arrows in the process referred to as $Draw$. Next, the visual condition $\mathcal{X}^i_{\text{vis}}$ is constructed through alpha compositing between the scene image $\mathcal{I}^s$ and the visual prompt.

\begin{equation}
\mathcal{X}^{i}_{\text{vis}} \;=\; (1 - \alpha) \cdot \mathcal{I}^{s} + \alpha \cdot \mathcal{P}^{i}
\end{equation}

Additionally, the past trajectory is represented as a heatmap, denoted as the history heatmap $\mathcal{X}^i_{\text{heatmap}}$. The heatmap is concatenated with the semantic map $\mathcal{X}^i_{\text{sem}}$ to form the semantic condition $\mathcal{X}^i_{\text{sem}}$ as follows:

\begin{equation}
\mathcal{X}^{i}_{\text{sem}} \;=\; \mathcal{X}^{i}_{\text{heatmap}} \oplus  \mathcal{I}^{s}_{\text{sem}}
\end{equation}
\noindent where $\oplus$ denotes channel-wise concatenation.

Finally, we compute the goal probability map $\hat{\mathcal{Y}}^i_{\text{goal}}$, from which the 2D coordinates of the goal are sampled. The $\hat{\mathcal{Y}}^i_{\text{goal}}$ is defined as:

\begin{equation}
\hat{\mathcal{Y}}^i_{\text{goal}} \;=\; \sigma \Bigg[ \zeta \Big( \phi \big( \mathcal{X}^{i}_{\text{vis}} \big),\; \mathcal{X}^{i}_{\text{sem}} \Big) \Bigg]
\end{equation}
\noindent Here, $\zeta$ represents the U-Net-based goal module~\cite{unet}, $\phi$ refers to the pretrained visual encoder, and $\sigma$ is the sigmoid function.
Similar to prior work~\cite{ynet, goal_sar}, we adopt a U-Net-based encoder-decoder architecture~\cite{unet} for the goal module. However, unlike the original designs, we modify the skip connections to concatenate the encoded features from both $\mathcal{X}^i_{\text{sem}}$ and $\mathcal{X}^i_{\text{vis}}$ before passing them to the decoder.

To predict pedestrian trajectories using LLMs (Figure~\ref{fig:2b}), prompt engineering is needed to convert the past trajectory $\mathcal{X}^i_t$ from numeric form into a natural language prompt. 
A tokenization process is then required to input this prompt into the LLM. 
For both prompt generation and tokenization, we adopted the method proposed by LMTraj~\cite{lmtrajectory}.
Unlike previous studies, we incorporate the CoT goal context as input for the LLM. 
Rather than merely predicting the goal, CoT extracts a goal prompt by leveraging past trajectory data to represent the pedestrian’s intended goal. 
The LLM then focuses solely on generating a realistic trajectory toward the goal without performing additional goal prediction.
The CoT-based approach allows the model to better understand underlying movement patterns, enabling more accurate and context-aware future trajectory predictions. 
By structuring the trajectory generation process around a well-defined goal, this method significantly enhances the realism of the predicted paths.

\subsection{User-guided Trajectory Generation}
Unlike existing LLM-based models such as SOTA LMTraj~\cite{lmtrajectory}, our approach enables comprehensive trajectory modeling, not limited to simple trajectory prediction. Notably, our method allows users to provide high-level guidance, such as direction or group behavior, without the need to specify exact 2D goal coordinates. This makes our model more interpretable and adaptable to user input. Figure~\ref{fig:control} illustrates examples of user-guided trajectory generation.

To achieve this, we adjust the goal logit map $\zeta \Big( \phi \big( \mathcal{X}^{i}_{\text{vis}} \big), \; \mathcal{X}^{i}_{\text{sem}} \Big)$ by incorporating a guidance function $\mathcal{G} \big( \mathcal{X}^{i}_{t} \big)$ as a regularization term, allowing for user-guided trajectory adjustments. The modified goal probability map is computed as follows:

\begin{equation}
\hat{\mathcal{Y}}_{\text{goal}}^i \;=\; \sigma \Bigg[ \zeta \Big( \phi \big( \mathcal{X}^{i}_{\text{vis}} \big), \; \mathcal{X}^{i}_{\text{sem}} \Big) + \lambda \cdot \mathcal{G} \big( \mathcal{X}^{i}_{t} \big) \Bigg]
\label{eq4}
\end{equation}

We define guidance functions for two scenarios: direction guidance and group behavior. For direction (Figure~\ref{fig:3b}), logits, derived by $\mathcal{G}_{\text{direction}} \big( \mathcal{X}^{i}_{t} \big) = \max{\Big(0,\; 1 - \frac{|\theta - \theta_p|}{\theta_{\text{max}}} \Big)}$, are added to regions rotated by a user-defined angle relative to the pedestrian's current heading. $\theta$ is the target angle for the pedestrian, $\theta_p$ is the angle between each position and the current position, and $\theta_{\text{max}}$ is the maximum angle for adding logits. For group behavior (Figure~\ref{fig:3c}), logits, derived by $\mathcal{G}_{\text{group}} \big( \mathcal{X}^{i}_{t} \big) = \max{\Big(0,\; 1 - \frac{d_p}{d_{\text{max}}} \Big)}$, are added around the predicted goals of neighboring pedestrian. $d_p$ represents the distance to the neighboring pedestrian's future position, and $d_{\text{max}}$ is the maximum distance. The level of influence from this guidance is controlled via the parameter $\lambda$, which adjusts the strength of the user input.

\begin{algorithm}[t]
    \caption{GUIDE-CoT Training Scheme}
    \label{alg}
\begin{flushleft}
    \textbf{(Step 1) Goal-oriented visual prompt training scheme} \\
    \textbf{INPUT:} Past trajectories of pedestrians $\mathcal{X}$, Future trajectories of pedestrians $\mathcal{Y}$, Scene image $\mathcal{I}^s$, Semantic map $\mathcal{I}^s_{\text{sem}}$, Goal module $\zeta$, Pretrained visual encoder $\phi$, Batch size $N$  \\
    \textbf{OUTPUT:} Trained goal module $\zeta$
\end{flushleft}
    \begin{tabbing}
    \hspace{1em} \= \hspace{1em} \= \hspace{1em} \= \kill
    \textbf{for} batch in $\mathcal{X}$ \textbf{do} \\
    \> \textbf{for} $i = 1$ to $N$ \textbf{do} \\
    \> \> $\mathcal{P}^i \;\gets\; Draw(\mathcal{X}^i)$ \\
    \> \> $\mathcal{X}^{i}_{\text{vis}} \;\gets\; (1 - \alpha) \cdot \mathcal{I}^{s} + \alpha \cdot \mathcal{P}^{i}$ \\
    \> \> $\mathcal{X}^i_{\text{sem}} \;\gets\; \mathcal{X}^i_{\text{heatmap}} \oplus \mathcal{I}^s_{\text{sem}}$ \\
    \> \> $\hat{\mathcal{Y}}^i_{\text{goal}} \;\gets\; \sigma \big[ \zeta \big( \phi \big( \mathcal{X}^{i}_{\text{vis}} \big),\; \mathcal{X}^{i}_{\text{sem}} \big) \big]$ \\
    \> \textbf{end for} \\
    \> $\mathcal{L}_{\text{goal}} \;\gets\; \text{BCE}(\hat{\mathcal{Y}}_{\text{goal}},\; \mathcal{Y}_{\text{goal}})$ \\
    \> Backprop and update $\zeta$ \\
    \textbf{end for} \\
    \textbf{Return:} Trained goal module $\zeta$
    \end{tabbing}
\begin{flushleft}
    \textbf{(Step 2) CoT LLM training scheme} \\
    \textbf{INPUT:} Past trajectories of pedestrians $\mathcal{X}$, Future trajectories of pedestrians $\mathcal{Y}$, LLM, Batch size $N$ \\
    \textbf{OUTPUT:} Trained LLM
\end{flushleft}
    \begin{tabbing}
    \hspace{1em} \= \hspace{1em} \= \hspace{1em} \= \kill
    \textbf{for} batch in $\mathcal{X}$ \textbf{do} \\
    \> \textbf{for} $i = 1$ to $N$ \textbf{do} \\
    \> \> CoT prompt $\gets$ from ground-truth $\mathcal{Y}^i$ \\
    \> \> $\hat{\mathcal{Y}}^i_{\text{text}} \gets \text{LLM}(\text{prompt})$ \\
    \> \textbf{end for} \\
    \> $\mathcal{L}_{\text{text}} \gets \text{CE}(\hat{\mathcal{Y}}_{\text{text}},\; \mathcal{Y}_{\text{text}})$ \\
    \> Backprop and update $\text{LLM}$ \\
    \textbf{end for} \\
    \textbf{Return:} Trained LLM
    \end{tabbing}
\end{algorithm}

\subsection{Training and Inference Scheme}
The proposed approaches are learned through the following loss functions. First, for a goal-oriented visual prompt (Figure~\ref{fig:2a}), model learns to align the predicted goal probability map $\hat{\mathcal{Y}}_{\text{goal}}$ with the ground truth probability map $\mathcal{Y}_{\text{goal}}$, which is generated from the final 2D coordinates $p_{t+\tau_{\text{pred}}}$ of the ground truth trajectory $\mathcal{Y}_t$.
The $\mathcal{L}_{\text{goal}}$ is computed using binary cross entropy ($BCE$) as follows: $\mathcal{L}_{\text{goal}} \;=\; BCE \left( \hat{\mathcal{Y}}_{\text{goal}},\; \mathcal{Y}_{\text{goal}} \right)$
Next, for the chain-of-thought (CoT) LLM for trajectory generation (Figure~\ref{fig:2b}), the predicted text-based trajectory $\hat{\mathcal{Y}}_{\text{text}}$ is compared with the ground truth trajectory $\mathcal{Y}_{\text{text}}$, represented in text form, using cross entropy ($CE$) to compute the $\mathcal{L}_{\text{text}}$. As a result, the $\mathcal{L}_{\text{text}}$ is expressed as follows: $\mathcal{L}_{\text{text}} \;=\; CE \Big( \hat{\mathcal{Y}}_{\text{text}},\; \mathcal{Y}_{\text{text}} \Big)$

During training, the LLM uses the ground truth chain-of-thought from the dataset instead of the one generated by the goal-oriented visual prompt. This ensures that the goal-oriented visual prompt and LLM are trained independently, without influencing each other. The overall training process of the goal-oriented visual prompt and the CoT LLM is shown in Algorithm~\ref{alg}. At inference time, the chain-of-thought sampled from the goal-oriented visual prompt  is passed to the LLM to predict the final trajectory.

\begin{figure}[t]
    \centering
    \includegraphics[width=.46\textwidth]{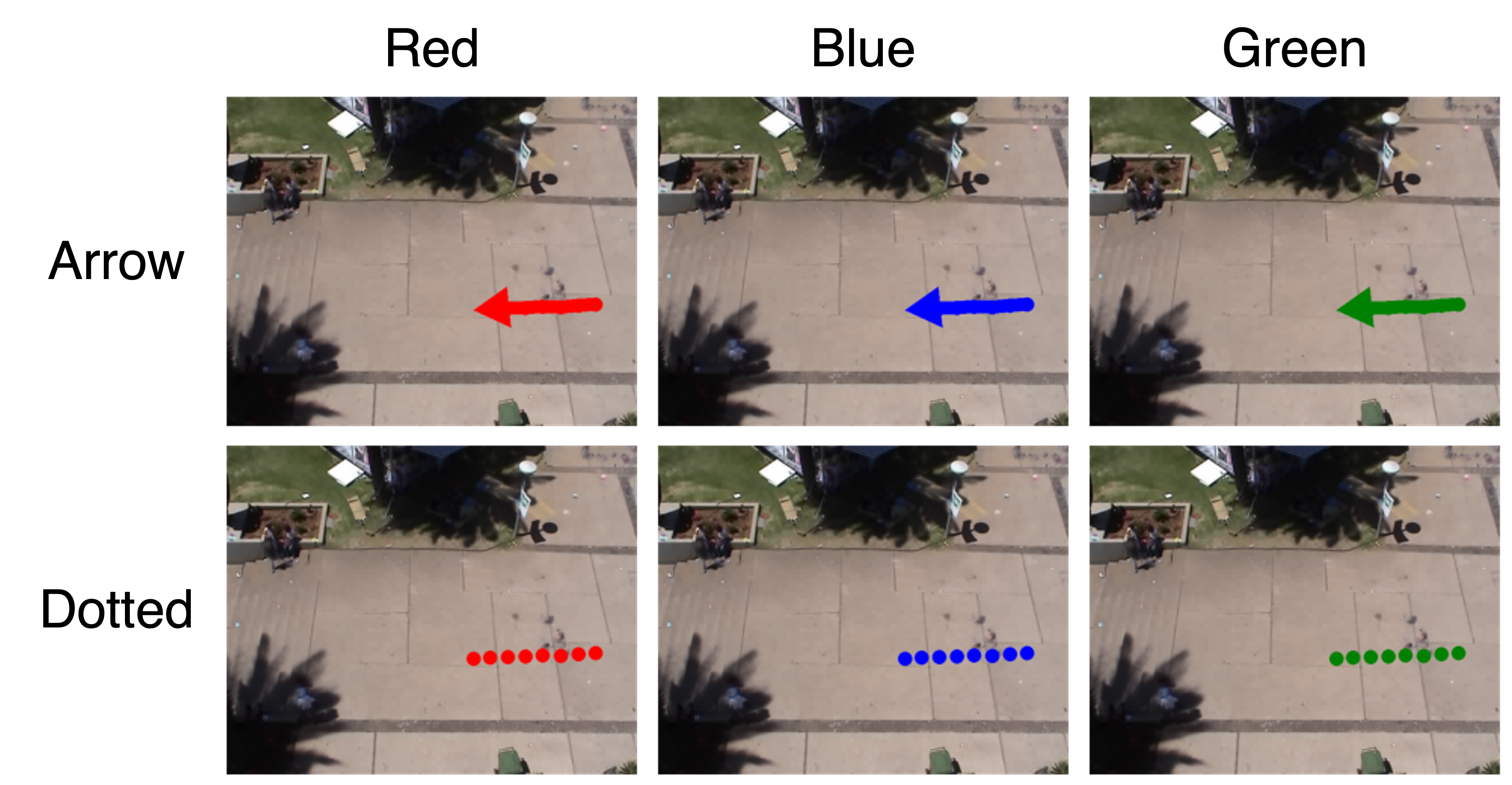}
     \caption{
     Examples of variants of our used visual prompts with different colors (i.e., red, blue, and green) and shapes (i.e., arrow and points).
     }
    \label{fig:visual_prompt}
\end{figure}

%% file: tex/4_experiments.tex
\section{Experimental Results}

\input{tables/tab1}
\input{tables/tab2}

\subsection{Implementation and Evaluation Details}\label{sec4a}
\myparagraph{Datasets.} 
We evaluated the trajectory prediction performance using the widely used ETH~\cite{eth} and UCY~\cite{ucy} datasets. Following prior works, we adopted the leave-one-out cross-validation setting for our experiments. We also followed the previous experimental settings~\cite{npsn, eigentrajectory, lmtrajectory}, i.e., dataset splits.

\myparagraph{Evaluation metrics.} For quantitative evaluation, we used well-known metrics: Average Displacement Error (ADE) and Final Displacement Error (FDE). The model predicted 20 possible future trajectories, from which the one with the smallest error was selected. All results were reported in meters.

\myparagraph{Implementation details.} For goal-oriented visual prompt, we employed the CLIP~\cite{clip} visual encoder based on the ResNet-50~\cite{resnet}. The goal module utilized a CNN-based U-Net encoder-decoder architecture. We trained the goal-oriented visual prompt for 50 epochs, with a batch size of 64, on a single RTX-3090Ti GPU, which took approximately 7 hours. For the LLM, we used the T5 small~\cite{t5} model, consistent with LMTraj~\cite{lmtrajectory}. The prompt used to transfer Goal CoT to the LLM is: “Pedestrian $i$ will arrive at coordinate $p^i_{t+\tau_{\text{pred}}}$ after the next $\tau_{\text{pred}}$ frames.”. The experimental code can be found on our GitHub at \url{https://github.com/ai-kmu/GUIDE-CoT}

\begin{figure*}[t]
    \centering
    \includegraphics[width=\textwidth]{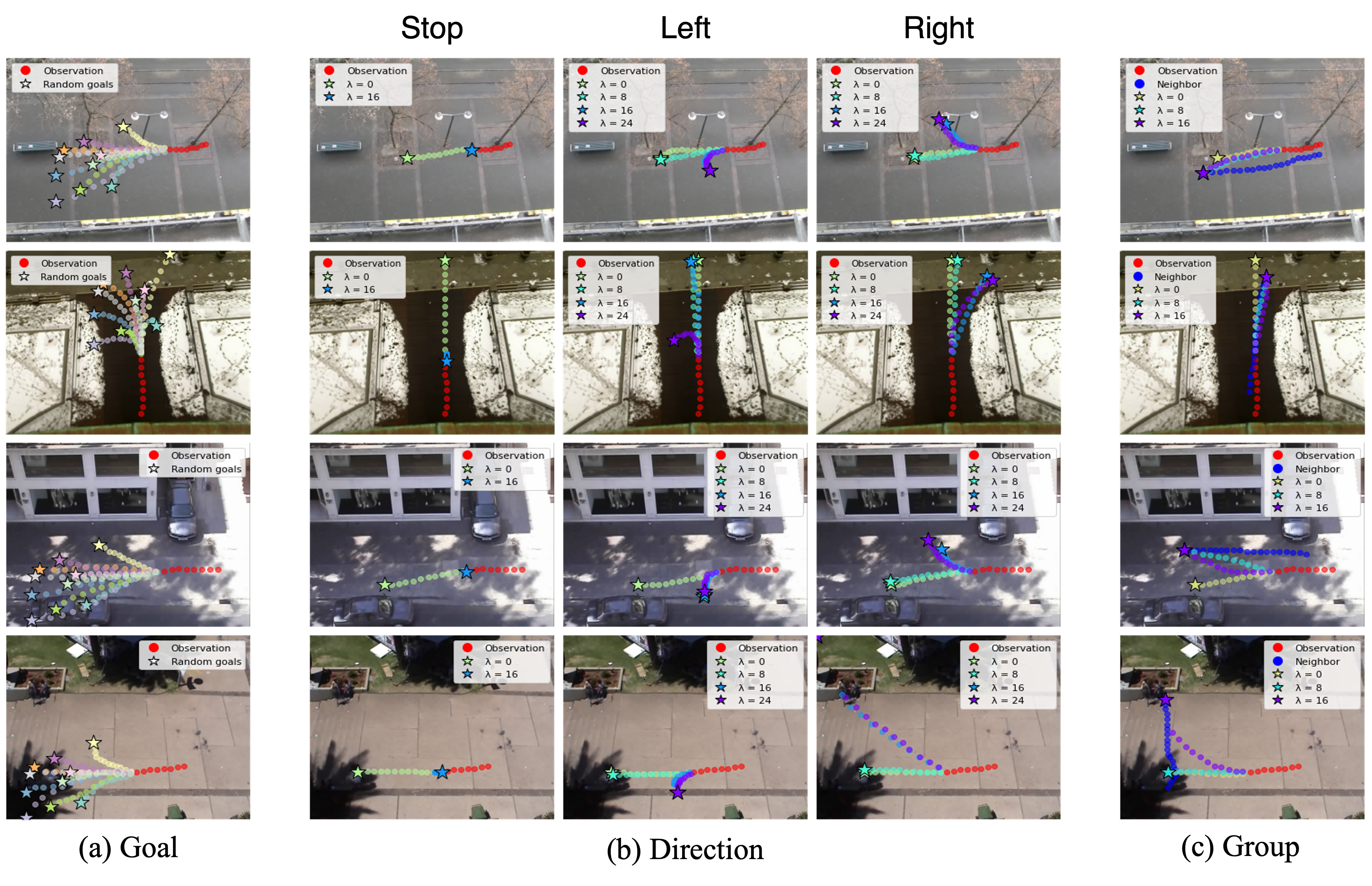}
    \captionsetup{skip=2pt} % 간격 조정
    \caption{
    Visualization of controllable trajectory generation based on user guidance. The red points represent the pedestrian’s observed trajectory, and the stars indicate the generated goals according to the goal-oriented visual prompt. Points in other colors show predicted trajectories from goal adjustments, illustrating variations like stopping, turning, and grouping with nearby pedestrians. These trajectories demonstrate the model’s adaptability to user-guided trajectory generation.
    }
    \label{fig:visualize}
\end{figure*}

\subsection{Trajectory Prediction Performance Comparison with Existing Approaches}\label{sec4b}
The results of pedestrian trajectory prediction on the ETH/UCY datasets are presented in Table~\ref{tab1}. 
The best performance is highlighted in bold, and the second-best is underlined. 
Our proposed method achieves ADE comparable to existing models and demonstrates the best performance in terms of average FDE across the datasets.
This demonstrates that the goal-oriented visual prompt effectively contributes to improving goal prediction performance.

\myparagraph{Comparison with LLM-based models.}
Our method surpasses LMTraj-SUP~\cite{lmtrajectory}, which also utilizes an LLM. 
This demonstrates that providing the LLM with appropriate context, rather than merely tasking it with inferring the full trajectory, enables more effective reasoning and leads to enhanced predictive performance.
Furthermore, the incorporation of goal-oriented visual prompts not only enhances trajectory estimation but also allows the model to leverage spatial and visual information more effectively.
This leads to a deeper understanding of pedestrian intent, resulting in more accurate predictions across diverse environments.

Our method provides flexibility through user-guided trajectory generation, allowing high-level guidance without precise goal coordinates. 
This adaptability allows precise handling of complex pedestrian behaviors and dynamic environments, outperforming previous models. 
Its ability to generalize across datasets demonstrates robustness, ensuring reliable performance under diverse conditions.
Integrating vision-based and semantic map-based goal prediction creates a synergistic effect, supporting more context-aware predictions. 
By leveraging multiple modalities instead of relying solely on historical data, our approach ensures realistic and adaptable trajectories suited to complex urban environments.
These strengths enable state-of-the-art performance while introducing new capabilities for user-informed trajectory generation.

\myparagraph{Comparison with goal-based models.}
GUIDE-CoT shares similarities with existing goal-based pedestrian trajectory prediction methods, as it predicts both the goal and the trajectory leading toward it. However, our method achieves superior performance over the representative goal-based approach, Goal-SAR~\cite{goal_sar}, in terms of goal prediction accuracy, as measured by FDE. Although both GUIDE-CoT and Goal-SAR~\cite{goal_sar} employ the same U-Net-based goal module~\cite{unet} to process semantic conditions, our results demonstrate that the introduction of goal-based visual prompts plays a crucial role in enhancing performance.

These performance improvements offer two key insights. First, incorporating well-designed visual prompts enables the pretrained visual encoder to extract more meaningful features. Second, these enriched features help the model effectively utilize environmental information for goal prediction, which significantly contributes to improved overall performance.

\subsection{Effect of Goal-oriented Visual Prompt}\label{sec4e}
We present the first approach to represent pedestrian trajectories as visual prompts for reasoning. To further analyze the effectiveness of our approach, we conduct experiments to evaluate the impact of different combinations of visual prompts and pretrained visual encoders on model performance.

\myparagraph{For visual prompts.}
As shown in Figure~\ref{fig:visual_prompt}, we test six types of visual prompts, combining three colors—red, blue, and green—with two shapes: arrows connecting the pedestrian’s starting and ending points, and dotted points representing the movement path. 
Most visual prompts yield performance comparable to leading methods, with the red arrow achieving the remarkable results as shown in Table~\ref{tab2a}.
These findings suggest that the CLIP~\cite{clip} visual encoder effectively interprets a wide range of visual prompts.

\myparagraph{For conditions.}
Table~\ref{tab2b} presents the performance evaluation from ablation experiments, which test the effect of using different input conditions for the goal module when generating the goal probability map. In the experiments where only the visual condition was applied—consisting of scene images and visual prompts—we combined a pretrained visual encoder with a CNN-based decoder, structured similarly to U-Net~\cite{unet}, and trained only the decoder for evaluation. The results indicate that the best performance was achieved when both conditions (visual and semantic) were applied together. This suggests that integrating features extracted by the pretrained visual encoder into existing goal modules~\cite{ynet, goal_sar} significantly improves the model’s ability to predict accurate goals.

\myparagraph{For pretraining strategies.}
Table~\ref{tab2c} presents a comparison of different visual encoder pretraining strategies. We evaluated the performance of the ImageNet-1K~\cite{imagenet}, CLIP~\cite{clip}, and Remote-CLIP~\cite{remoteclip} visual encoders. The results show that CLIP consistently outperforms ImageNet-1K across all datasets, highlighting its effectiveness in extracting relevant features for trajectory prediction. This superior performance can be attributed to CLIP’s ability to leverage multimodal data and align visual features with contextual information more effectively than traditional pretraining methods.

As a result, leveraging visual prompts along with a pretrained visual encoder has a clear positive impact on goal prediction by enhancing the model’s contextual understanding of pedestrian intent. Our findings align with prior research using visual prompts for reasoning tasks~\cite{clip_red_circle, vip-llava}, indicating that goal prediction shares key similarities with general reasoning tasks. This suggests that advancements in reasoning research can further improve goal prediction performance.

\subsection{Effect of User-guided Trajectory Generation}\label{sec4c}
We perform a qualitative analysis to evaluate user-guided trajectory generation using three types of high-level guidance: goal, direction, and group. The visual results are shown in Figure~\ref{fig:visualize}.
For goal guidance, a random goal was assigned for the pedestrian. For direction guidance, we tested three scenarios: stop, left, and right. For group guidance, a random neighboring pedestrian was selected, and the model was guided to group the target pedestrian with them. In Figure~\ref{fig:visualize}a, the randomly assigned goal led the LLM to generate realistic trajectories that followed the given goal accurately.

For direction guidance (Figure~\ref{fig:visualize}b), we tested three commands: stop, left, and right. Without guidance ($\lambda=0$), the predicted trajectory followed a straight path to the nearest goal. As $\lambda$ increased, the trajectory adjusted to the specified direction. Importantly, the model avoided assigning goals to non-traversable areas like obstacles or off-road locations. This behavior stems from our method, where the guidance function $\mathcal{G} \big( \mathcal{X}^{i}_{t} \big)$ modifies the goal logit map, fine-tuning the path rather than replacing it.

For group guidance (Figure~\ref{fig:visualize}c), we selected a random neighboring pedestrian and guided the target pedestrian to form a group with them. When $\lambda=0$, the red and blue pedestrians were predicted to remain independent. However, as $\lambda$ increased, the trajectories adjusted to bring them closer, successfully forming a group.

%% file: tables/tab1.tex
\begin{table*}[!t]
\centering
\renewcommand{\arraystretch}{1.3}  
\caption{Comparison of ADE/FDE across different models on the ETH/UCY datasets. The 1st/2nd best performances are indicated in boldface and underline. $\dagger$: Each model is evaluated under the common pedestrian trajectory prediction setting, i.e., consistent dataset splits and evaluation units (meters). $\ddagger$: Results for these models are reproduced.
% \jinkyu{can we add two additional columns to denote whether the given method is goal-based models or LLM-based models or neither.}
}
\label{tab1}
\resizebox{\textwidth}{!}{
\begin{tabular}{l|cccccc}
\toprule
%\multirow{3}{*}{\textbf{Model}} & \multicolumn{6}{c}{\large \textbf{Dataset}} \\ \cmidrule{2-7}
\multirow{2}{*}{\textbf{Model}} & \textbf{ETH}         & \textbf{HOTEL}       & \textbf{UNIV}        & \textbf{ZARA1}       & \textbf{ZARA2}       & \textbf{AVG}         \\
 & \footnotesize ADE (↓) / FDE (↓) & \footnotesize ADE (↓) / FDE (↓) & \footnotesize ADE (↓) / FDE (↓) &  \footnotesize ADE (↓) /FDE (↓) & \footnotesize ADE (↓) / FDE (↓) & \footnotesize ADE (↓) / FDE (↓) \\
\midrule
Social-GAN (18' CVPR)~\cite{social_gan}      & 0.77 / 1.40          & 0.43 / 0.88          & 0.75 / 1.50          & 0.35 / 0.69          & 0.36 / 0.72          & 0.53 / 1.04          \\
PECNet (20' ECCV)$^{1\dagger}$~\cite{pecnet}             & 0.61 / 1.07          & 0.22 / 0.39          & 0.34 / 0.56          & 0.25 / 0.45          & 0.19 / 0.33          & 0.32 / 0.56          \\
Trajectron++ (20' ECCV)$^\dagger$~\cite{trajectron++}    & 0.61 / 1.03          & 0.20 / 0.28          & 0.30 / 0.55          & 0.24 / 0.41          & 0.18 / 0.32          & 0.31 / 0.52          \\
AgentFormer (21' ICCV)$^\dagger$~\cite{agentformer}       & 0.46 / 0.80          & 0.14 / 0.22          & 0.25 / 0.45          & \underline{0.18} / 0.30          & \underline{0.14} / 0.24          & 0.23 / 0.40          \\
MID (22' CVPR)$^\dagger$~\cite{mid}              & 0.57 / 0.93          & 0.21 / 0.33          & 0.29 / 0.55          & 0.28 / 0.50          & 0.20 / 0.37          & 0.31 / 0.54          \\
Goal-SAR (22' CVPRW)$^{1\dagger\ddagger}$~\cite{goal_sar}          & 0.60 / 0.81 & 0.19 / 0.18 & 0.85 / 0.59 & 0.20 / 0.30 & 0.21 / 0.27 & 0.41 / 0.43 \\
NPSN (22' CVPR)~\cite{npsn}             & \textbf{0.36} / 0.59          & 0.16 / 0.25          & 0.23 / 0.39          & \underline{0.18} / 0.32          & \underline{0.14} / 0.25          & \textbf{0.21} / 0.36          \\
SocialVAE (22' ECCV)~\cite{socialvae}        & 0.41 / 0.58          & 0.13 / 0.19          & \textbf{0.21} / \underline{0.36}          & \textbf{0.17} / \underline{0.29}          & \textbf{0.13} / \underline{0.22}          & \textbf{0.21} / 0.33          \\
EqMotion (23' CVPR)~\cite{eqmotion}        & 0.40 / 0.61          & \underline{0.12} / 0.18          & 0.23 / 0.43          & \underline{0.18} / 0.32          & \textbf{0.13} / 0.23          & \textbf{0.21} / 0.35          \\
EigenTrajectory (23' ICCV)~\cite{eigentrajectory}  & \textbf{0.36} / 0.53          & \underline{0.12} / 0.19          & 0.24 / 0.43          & 0.19 / 0.33          & \underline{0.14} / 0.24          & \textbf{0.21} / 0.34          \\
LED (23' CVPR)~\cite{led}            & 0.39 / 0.58          & \textbf{0.11} / 0.17          & 0.26 / 0.43          & \underline{0.18} / \textbf{0.26}          & \textbf{0.13} / \underline{0.22}          & \textbf{0.21} / 0.33          \\
LMTraj-SUP (24' CVPR)$^{2}$~\cite{lmtrajectory}     & 0.41 / \underline{0.51}          & \underline{0.12} / \underline{0.16}          & \underline{0.22} / \textbf{0.34}          & 0.20 / 0.32          & 0.17 / 0.27          & \underline{0.22} / \underline{0.32}          \\ 
\noalign{\vskip 2pt}
\hline
\noalign{\vskip 2pt}
GUIDE-CoT (Ours)$^{1,2}$             & \underline{0.38} / \textbf{0.43} & 0.13 / \textbf{0.15}          & 0.34 / 0.48          & 0.19 / \underline{0.29}          & 0.16 / \textbf{0.21}          & 0.24 / \textbf{0.31}       \\ \bottomrule
\end{tabular}}
% \begin{tablenotes}[flushleft]
%     \item \footnotesize 1: Goal-based methods. 2: LLM-based methods
% \end{tablenotes}
\par\raggedright\footnotesize{1: Goal-based methods. 2: LLM-based methods}
\end{table*}

%% file: tables/tab2.tex
\begin{table}[]
    \centering
    \caption{
    Ablation study with (a) different types of visual prompts, (b) different combinations of visual conditions, and (c) different pretraining strategies (with the same ResNet-50~\cite{resnet} backbone). We use ETH/UCY datasets and report FDE scores (lower is better).
    %Quantitative analysis of goal-oriented visual prompt on ETH/UCY datasets in terms of FDE (lower is better).
    }
    \label{tab2}
    \subcaption[]{Effect on types of visual prompts}
    \label{tab2a}
    {\setlength{\tabcolsep}{2.5pt}
    \begin{tabular}{l|ccccc|c}
        \toprule
        & \textbf{ETH} & \textbf{HOTEL} & \textbf{UNIV} & \textbf{ZARA1} & \textbf{ZARA2} & \textbf{Avg.} \\
        \midrule
        \bluetext{Blue} points & 0.46 & 0.17 & 0.50 & \textbf{0.28}  & 0.22 & 0.33 \\
        \bluetext{Blue} arrow & 0.46 & 0.17 & 0.50 & \textbf{0.28} & 0.22 & 0.33 \\
        \midrule
        \greentext{Green} points & \textbf{0.43} & 0.16 & 0.50 & 0.29 & \textbf{0.21} & 0.32 \\
        \greentext{Green} arrow & 0.44 & 0.16 & 0.50 & \textbf{0.28}  & 0.22 & 0.32 \\
        \midrule
        \redtext{Red} points & 0.46 & 0.17 & 0.50 & \textbf{0.28} & 0.22 & 0.33 \\
        \redtext{Red} arrow & \textbf{0.43} & \textbf{0.15} & \textbf{0.48} & 0.29 & \textbf{0.21} & \textbf{0.31} \\
    \bottomrule
    \end{tabular}
    }
% \vspace{1em}
\subcaption[]{Effect on combinations of visual prompts}
\label{tab2b}
{\setlength{\tabcolsep}{3.2pt}
\begin{tabular}{cc|ccccc|c}
\toprule
$\mathcal{X}_{\text{sem}}$ & $\mathcal{X}_{\text{vis}}$ & \textbf{ETH} & \textbf{HOTEL} & \textbf{UNIV} & \textbf{ZARA1} & \textbf{ZARA2} & \textbf{Avg.} \\
\midrule
\checkmark & & 0.66 & 0.22 & 0.64 & 0.36 & 0.24 & 0.42 \\
 & \checkmark & 0.66 & 0.32 & 1.22 & 0.60 & 0.33 & 0.51 \\
\checkmark & \checkmark &  \textbf{0.43}& \textbf{0.15} &\textbf{0.48}& \textbf{0.29} & \textbf{0.21} & \textbf{0.31} \\
\bottomrule
\end{tabular}
}
% \vspace{1em}
\subcaption[]{Effect on pretraining strategies}
\label{tab2c}
{\setlength{\tabcolsep}{1.2pt}
\begin{tabular}{l|ccccc|c}
\toprule
 & \textbf{ETH} & \textbf{HOTEL} & \textbf{UNIV} & \textbf{ZARA1} & \textbf{ZARA2} & \textbf{Avg.} \\
\midrule
ImageNet-1K~\cite{imagenet} & 0.44 & 0.18 & 0.49 & 0.30 & 0.22 & 0.33 \\
Remote-CLIP~\cite{remoteclip} & \textbf{0.41} & 0.17 & 0.49 & \textbf{0.28} & 0.22 & \textbf{0.31}  \\
CLIP~\cite{clip} & 0.43& \textbf{0.15} &\textbf{0.48}& 0.29 & \textbf{0.21} & \textbf{0.31} \\
\bottomrule
\end{tabular}
}
\end{table}

% \begin{table}
% \caption{Comparison of goal-oriented visual prompts based on different types of visual prompts on ETH/UCY datasets and FDE($\downarrow$) as the evaluation metric.}
% \label{tab2}
% {\setlength{\tabcolsep}{2.5pt}
% \begin{tabular}{c|cccccc}
% \toprule
%  & \textbf{ETH} & \textbf{HOTEL} & \textbf{UNIV} & \textbf{ZARA1} & \textbf{ZARA2} & \textbf{AVG} \\
% \midrule
% Red arrow & \textbf{0.43} & \textbf{0.15} & \textbf{0.48} & 0.29 & \textbf{0.21} & \textbf{0.31} \\
% Red dotted & 0.46 & 0.17 & 0.50 & \textbf{0.28} & 0.22 & 0.33 \\
% Blue arrow & 0.46 & 0.17 & 0.50 & \textbf{0.28} & 0.22 & 0.33 \\
% Blue dotted & 0.46 & 0.17 & 0.50 & \textbf{0.28}  & 0.22 & 0.33 \\
% Green arrow & 0.44 & 0.16 & 0.50 & \textbf{0.28}  & 0.22 & 0.32 \\
% Green dotted & \textbf{0.43} & 0.16 & 0.50 & 0.29 & \textbf{0.21} & 0.32 \\
% \bottomrule
% \end{tabular}
% }
% \end{table}

%% file: tex/5_conclusion.tex
\section{Conclusion}
We propose GUIDE-CoT, a novel framework that enhances pedestrian trajectory prediction by integrating goal-oriented visual prompts with a chain-of-thought LLM. Our method leverages visual prompts and pretrained visual encoders to deliver precise goal context to the LLM, improving both prediction accuracy and adaptability. Unlike existing LLM-based trajectory prediction methods, GUIDE-CoT uses the goal context as an input, enabling controllable trajectory generation by adjusting the goal dynamically. Extensive experiments on the ETH/UCY datasets demonstrate the effectiveness of our approach, showing that GUIDE-CoT not only outperforms current methods but also introduces new capabilities for user-guided, context-aware trajectory predictions.

One key limitation of this study is that the effectiveness of visual information can vary with the environment. In areas with distinct obstacles, such as trees and buildings, pedestrian future goals can be predicted with relatively high accuracy. However, in open spaces like UNIV, where there are few obstacles, predicting these goals becomes more challenging (see Table~\ref{tab1}). To address this issue, future work should explore integrating additional contextual cues from the surrounding environment--beyond just visual information--into the LLM.